\documentclass[10pt,twocolumn,letterpaper]{article}
\usepackage{wacv}
\usepackage{times}
\usepackage{epsfig}
\usepackage{graphicx}
\usepackage{amsmath}
\usepackage{amssymb}
\usepackage{booktabs}
\newcommand\blfootnote[1]{%
  \begingroup
  \renewcommand\thefootnote{}\footnotetext{#1}%
  \addtocounter{footnote}{+1}%
  \endgroup
}

\usepackage{color}

\usepackage{algorithm}
\usepackage[noend]{algpseudocode}
\usepackage{xcolor}
\algdef{SE}[SUBALG]{Indent}{EndIndent}{}{\algorithmicend\ }%
\algtext*{Indent}
\algtext*{EndIndent}

\def\wacvPaperID{1614} 

\wacvapplicationstrack 

\wacvfinalcopy 

\ifwacvfinal
\usepackage[breaklinks=true,bookmarks=false]{hyperref}
\else
\usepackage[pagebackref=true,breaklinks=true,colorlinks,bookmarks=false]{hyperref}
\fi

\begin{document}

\title{Are Straight-Through gradients and Soft-Thresholding\\ all you need for Sparse Training?}

\author{Antoine Vanderschueren$^1$\\
UCLouvain, Belgium\\
{\tt\small antoine.vanderschueren@uclouvain.be}
\and
Christophe De Vleeschouwer$^1$\\
UCLouvain, Belgium\\
{\tt\small christophe.devleeschouwer@uclouvain.be}
}

\maketitle\blfootnote{\hspace{-0.5em}$^1$Part of this work has been funded by the Walloon Region project SmartGate N°1910087, and by the Fonds de la Recherche Scientifique – FNRS}
\thispagestyle{empty}

\begin{abstract}
Turning the weights to zero when training a neural network helps in reducing the computational complexity at inference. 
To {\it progressively increase} the sparsity ratio in the network {\it without causing sharp weight discontinuities} during training, our work combines soft-thresholding and straight-through gradient estimation to {\it update the raw, i.e. non-thresholded, version} of zeroed weights.
Our method, named \textbf{ST-3} for straight-through/soft-thresholding/sparse-training\footnote{Source code and weights available at \url{https://github.com/vanderschuea/stthree}
}, obtains SoA results, both in terms of accuracy/sparsity and accuracy/FLOPS trade-offs, when progressively increasing the sparsity ratio in a single training cycle. In particular, despite its simplicity, ST-3 favorably compares to the most recent methods, adopting differentiable formulations~\cite{Zhou2021} or bio-inspired neuroregeneration principles~\cite{Liu2021}. This suggests that the key ingredients for effective sparsification primarily lie in the ability to give the weights the freedom to evolve smoothly across the zero state while progressively increasing the sparsity ratio.
\end{abstract}

\section{Introduction}

State-of-the-art neural networks are composed of a few million parameters, leading to a few billion computations per inference. To limit these computations, sparse networks have been thoroughly investigated in the past few years \cite{Zhu2018,Frankle2019,Liu2019,Zhou2019a,Lee2019,Wortsman2019,Evci2020,Frankle2020,Renda2020,Li2020,Zhou2021,Liu2021}, and significant efforts have been devoted to their efficient hardware implementation \cite{Gray2017,Mishra2021}. Sparse networks reduce the \emph{inference complexity} by setting a majority of their weight parameters to zero. 
Straightforward approaches to train sparse networks are based on pruning \cite{Zhu2018,Han2015,Frankle2019}, meaning links are removed from the network (which is equivalent to setting their weight to zero)  with no chance of being re-activated later along the training. 
Since the pruning decision is generally based on weight magnitudes, this approach penalizes the weights that have to \emph{change signs} during training, since those weights end up being pruned when they cross the zero border \cite{Sanh2020}. 
In contrast, modern sparsification methods favor the emergence of zero weights, without definitely canceling the corresponding links,
allowing weights to freely switch between an active and inactive state. 
The wide variety of approaches that have been proposed for this purpose are surveyed in Section~\ref{sec:sota}. 
Our work proposes a conceptually and computationally simple solution that (surprisingly) quite favorably compares to this rich array of prior art. 

Our proposal mitigates computational complexity at training by restricting the learning to a single round of gradient descent iterations. This is in contrast with the iterative post-training pruning solutions~\cite{Frankle2020,Renda2020}, which achieve SoA accuracy/sparsity trade-off, but suffer from an extremely large computation cost due to the multiple rounds of training required to progressively increase the pruning ratio.

As many previous works building on a single round of training, our method adopts a dynamic weight thresholding procedure to progressively increase the ratio of weights that are set to zero along the training, and to balance zeroed weights across layers. 
Our work is however specific in that (i) it continuously updates zeroed weights all along the training, and (ii) it sets the weights to zero based on soft-thresholding, instead of hard-thresholding, popular in previous works. Those specificities are detailed as follows. 

To continuously update the zeroed weights, our method takes inspiration from the training of quantized neural networks (QNN), where the raw (unquantized) weights are updated with a Straight-through-estimator (STE) to avoid a gradient that goes to zero due to the step-function of the quantizer \cite{Hubara2016,Li2016}. We apply this same principle to the weights that are 'quantized' to zero during network sparsification. This means that raw and non-thresholded weights are maintained and updated in the backward path, even if their thresholded version are zeroed in the forward path.
STE allows gradients and momentum to switch positive weights towards negative values (and vice-versa), without getting stuck in an inactive state when being close to zero.

To turn low magnitude weights to zero, thereby sparsifying the network, soft-thresholding is adopted. This prevents the abrupt changes of forward-path weight values induced by hard-thresholding operators, which may turn a small gradient update into a sharp weight discontinuity. Such abrupt discontinuities hamper the network accuracy, especially at high sparsity rates, where they lead to premature layer collapse \cite{Tanaka2020}, as illustrated in Figure \ref{fig:ste}.

Despite its simplicity, our approach surpasses the accuracy/sparsity trade-offs achieved by the most recent (and generally more sophisticated) single training cycle alternatives. Thereby, it reveals that the key ingredients required to train effective sparse networks primary consist in (1) giving the weights the capability to evolve freely and smoothly between active and inactive states during training, especially in the early epochs; 
and (2) progressively increasing the sparsity ratio without causing sharp discontinuities in weight values along the training.  

Another convincing experimental argument in favor of our method lies in the fact that, when combined with iterative and thus quite complex (due to multiple rounds of training cycles) post-training pruning solutions proposed in ~\cite{Frankle2020,Renda2020}, our ST-3 defines a novel SoA accuracy/sparsity trade-off. 
Overall, due to its effectiveness but also computational and conceptual simplicity, our ST-3 provides a novel and valuable baseline to assess future sparsification methods. Despite the key ingredients of ST-3 having been independently proposed in earlier works (generally to implement multi-cycle pruning or quantization), our work is novel and original in the way it integrates them in a method specifically suited to single-cycle training of sparse networks. Our work is also the first one to demonstrate that those conceptually simple ingredients are sufficient to reach SoA performance, when properly combined.

\section{Related Works}
\label{sec:sota}
While zeroing the weights in a structured manner might lead to higher acceleration when considering specific hardware implementations, the accuracy obtained with unstructured sparsity is generally higher at a given sparsity level\cite{Li2017,He2019a,Wang2020},
and significant speed up can be achieved with unstructured sparsity on off-the-shelf hardware \cite{Kalchbrenner2018,Park2017,Gale2019,Mishra2021}. Hence, this paper concentrates on \emph{unstructured} sparsity. 
Initial approaches to learn sparse networks simply removed links from the network at the end of an entire training cycle, based on their respective L1 magnitude \cite{LeCun1990,Han2015}. This however comes at the cost of a substantial loss in prediction accuracy compared to a dense network, including when the remaining \emph{active} weights undergo a \emph{finetuning} step (i.e. a shorter training cycle, with small learning rate).
Modern approaches do not consider the definitive and abrupt removal of links. They instead implement partial pruning at the end of multiple training cycles or design solutions for progressive weight zeroing.

Reference approaches to progressively increase the zeroed weights ratio have been studied in \cite{Zhu2018} and \cite{Kusupati2020}. Global Magnitude Pruning (GMP) \cite{Zhu2018} progressively prunes the links along the training. A related method, named STR, sets the weights that lie below a certain threshold to zero, using soft-thresholding. In contrast to our ST-3, only the non-zeroed weights are updated by gradients. 

A few previous methods have revealed the benefit of a {\it non-permanent} zeroing of weights. RIGL \cite{Evci2020} builds on gradient momentum to resurrect some of the zeroed weights. GraNet \cite{Liu2021} boosts the pruning plasticity through bio-inspired neuroregeneration principles. ProbMask \cite{Zhou2021} learns a probabilistic pruning mask that is transformed into a binary one by sampling (multiple times, each requiring gradient computations) a Gumbel distribution, thus making it trainable via gradient descent. 
Our ST-3 is conceptually and computationally simpler since it relies on straight-through gradient estimation to continuously update zeroed weights. Our experiments also reveal that when combined with soft-thresholding, it reaches better accuracy-sparsity trade-offs than earlier works.

Beyond its specificities regarding weight zeroing (with soft-thresholding) and zeroed weight updates (based on straight-through estimation, as done when training quantized networks \cite{Hubara2016,Li2016}), our ST-3 adopts conventional solutions regarding three other important design questions : 

{\bf Weight significance definition.} Most methods use gradient-magnitude \cite{Evci2020}, weight-magnitude \cite{Han2015,Zhu2018,Kusupati2020}, or a mix of both \cite{Tanaka2020} to select the weights to set to zero after some preliminary training. As an alternative, DNW \cite{Wortsman2019} uses neural architecture search to discover neural wirings, i.e. independent critical connections between channels, thereby finding a sparse subnetwork of the complete graph. Our work adopts the main trend, zeroing weights based on their magnitude. 

{\bf Distributing sparsity across layers.}   
It has been widely documented \cite{Mocanu2018,Evci2020,Lee2021} that the network prediction accuracy is more impacted when pruning occurs close to the input. In a CNN, it is however more beneficial from a computational point of view to prune the first layers since they correspond to high-resolution channels.
Two heuristics have been recently proposed to address this trade-off.
First, Erdos-Rényi-Kernel (ERK) \cite{Mocanu2018,Evci2020} proposes to scale the global sparsity ratio (= ratio of zeros to the total parameter count) with a layer-wise factor. 
so as to induce a higher (smaller) sparsity for layers with more (less) parameters, e.g. the first convolution layers of a ResNet-50 remain dense, thereby preserving accuracy. 
Second, Layer-Aware-Pruning (LAMP) \cite{Lee2021} selects the weights to prune globally, but relies on a score that is assigned to each weight rather than on its magnitude. 
This score is computed on the sorted and flattened array of weights ($\mathbf{W^l}_\text{sorted}$) of each layer ($l$). 

Both ERK and LAMP target higher accuracy at the cost of an increased complexity (as measured in FLOPS). However, our experiments did not reveal significant accuracy gains when using one of those heuristics. Hence, unless said otherwise, weights are considered independently of their layer in our experiment. As detailed in Section 3, to handle use cases where the gain in number of operations (FLOPS) is more important than the accuracy, we propose to normalize the weights by the size of their kernel, which approximates the standard deviation of the weights in the layer kernel. Consequently, this normalization reduces the risk to set a whole kernel to zero, thereby preserving accuracy at high sparsity ratio.

{\bf Controlling the global level of sparsity along the training.} This third question is the one that most impacts the training computational cost. It is therefore central to our paper, which targets simplicity without penalizing accuracy. 
Most authors \cite{Frankle2020,Frankle2020b,Renda2020,Leclerc2020} agree on the fact that: (i) the pruning should not start at the beginning of training, due to the \emph{heavy changes} a model undergoes in the early stage of training; and (ii) the increase of sparsity should be progressive, to limit the interference between weight update and weight zeroing. Two strategies co-exist to progressively increase the sparsity ratio.

The first one recommends decoupling training iterations and weight zeroing, to promote training consistency and thus convergence quality. Therefore, it builds on multiple training cycles, and is thus intrinsically more complex.
It increments the amount of zero weights by a constant fraction of non-zeroed weights at the end of each full training cycle. The learning rate - and optionally the weights' values - are rewinded (the acronym LRR is used to denote this rewinding process), i.e. reset, to an anterior value, before the training resumes. Doing so, SoA accuracy are obtained, but the training cost largely grows with the sparsity ratio~\cite{Frankle2020,Renda2020}. When adopting the same post-training sparsity increase than LRR, and thus iterating over multiple training cycles, our method outperforms the original LRR method, and defines a novel accuracy/sparsity SoA.

The second strategy primarily targets low computational load, and therefore increases the sparsity ratio along a single training cycle. GMP \cite{Zhu2018} proposes a cubic increase of the sparsity ratio during training, and has been adopted by ProbMask \cite{Zhou2021} and GraNet \cite{Liu2021}. This increase starts after a set amount of epochs, and finishes before the loss, and thus the gradients, become too small. 
Among the methods running in a single training cycle, ProbMask \cite{Zhou2021} and GraNet \cite{Liu2021} achieve highest accuracies. As mentioned above, \cite{Zhou2021}, transforms the discrete optimization problem associated to the pruning mask definition into a constrained expected loss minimization problem over a continuous probability space, which is solved using the Gumbel-Softmax trick. This however comes at the cost of multiple computations of gradients in each training iteration, and our experiments demonstrate that the accuracy in \cite{Zhou2021} fall behind our approach.
GraNet \cite{Liu2021}, and other similar approaches like \cite{Evci2020,liu2021we}, update the pruning mask every few thousands iterations, by regenerating connections when pruning others. This plasticity can be seen as a discrete approximation of the regeneration of weights that naturally occurs in our method when adopting soft-thresholding on top of the gradient accumulation inherent to our recommended straight-through estimator.

\section{Our Method}
\label{sec:ours}
This section introduces ST-3 as a new baseline for sparse training. 
It provides a relevant baseline candidate, because it is easy to implement (see Supplementary File), has very little overhead compared to regular dense training, and performs well with the default dense training parameters. ST-3 adopts weight magnitude soft-thresholding with threshold growing along training to progressively increase the sparsity ratio in the forward path. To preserve the opportunity for a zeroed weight to become significant again, similarly to what is done for quantized network training, it considers a straight-through gradient estimator to update a non-thresholded version of each weight maintained in the backward path. Improved training stability is achieved by scaling the weights in the forward path to compensate for the loss of magnitude inherent to soft-thresholding. Despite ST-3 builds on a set of existing or straightforward solutions (progressive increase of sparsity ratio with magnitude loss compensation, straight-through estimator, and soft-thresholding), it is novel in the sense that no previous work has combined all the ST-3 ingredients in a sparse training context. Our experimental study demonstrates the importance of including all elements to preserve ST-3 performance. They are detailed below, and all contribute to a stable and consistent training, aiming at minimizing the impact of weight zeroing on the gradient descent updates.

\paragraph{Straight-Through-estimation}
To control the sparsity ratio, our method sets to zero the weights that lie below some adaptive threshold. As mentioned before, a main requirement in sparse training is that the weights that are zeroed at some point during the training, because they are below the threshold, get the opportunity to become significant again in subsequent training iterations. Therefore, our method maintains a dense and continuously updated version of the weights in parallel to their sparse thresholded version, to be considered in the forward path but also to back-propagate the gradients. 
Following \cite{Bengio13,Penghang2019}, a straight-through estimator is considered to define the gradient with respect to the dense weight as a copy of the gradient with respect to the output of the thresholding operator, namely with respect to the sparse weight.
This differentiation of raw (to update) and thresholded (to use in forward and backward paths) weights is standard in the training of quantized neural networks \cite{Hubara2016,Mellempudi2017}, but is not common in the pruning community.

\begin{figure}[t]
  \centering
   \includegraphics[width=0.7\linewidth]{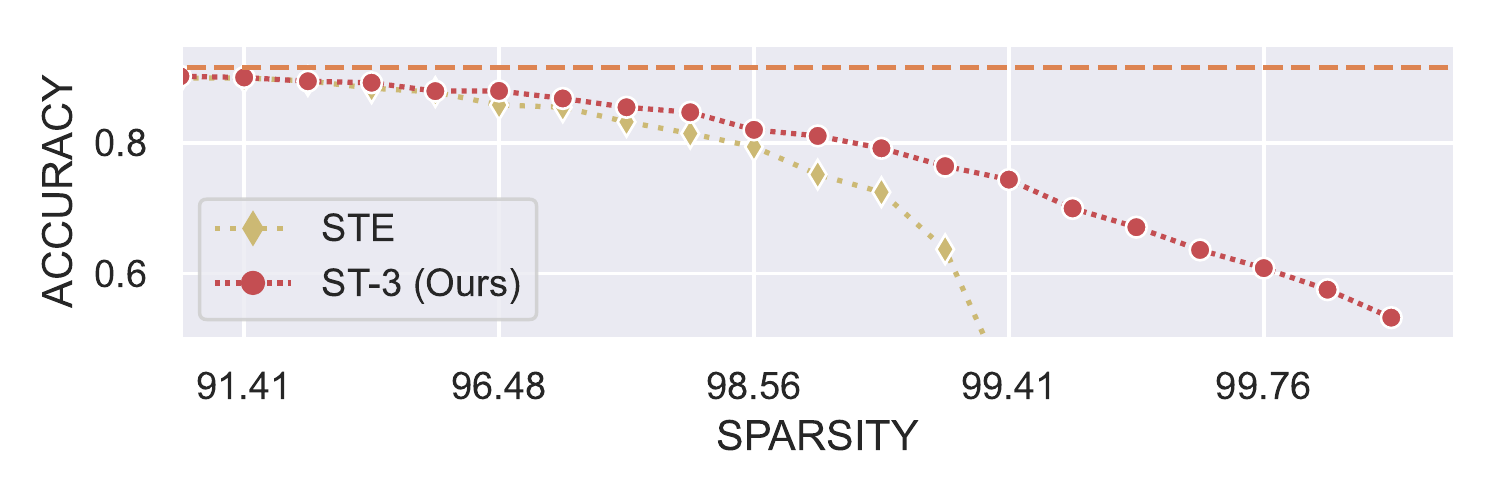}
   \caption{Accuracy of a sparse ResNet-20 on Cifar-10. Conventional straight-through-estimation (STE), with hard-thresholding,  fails at 99\% sparsity, as already pointed in \cite{Tanaka2020}. Replacing hard- by soft-thresholding reduces the gradient mismatch and aleviates the issue (Dense accuracy of 91.2\% orange dashed line)}
   \label{fig:ste}
\end{figure}

\paragraph{Soft Thresholding.}
\begin{figure}[t]
  \centering
   \includegraphics[width=0.7\linewidth]{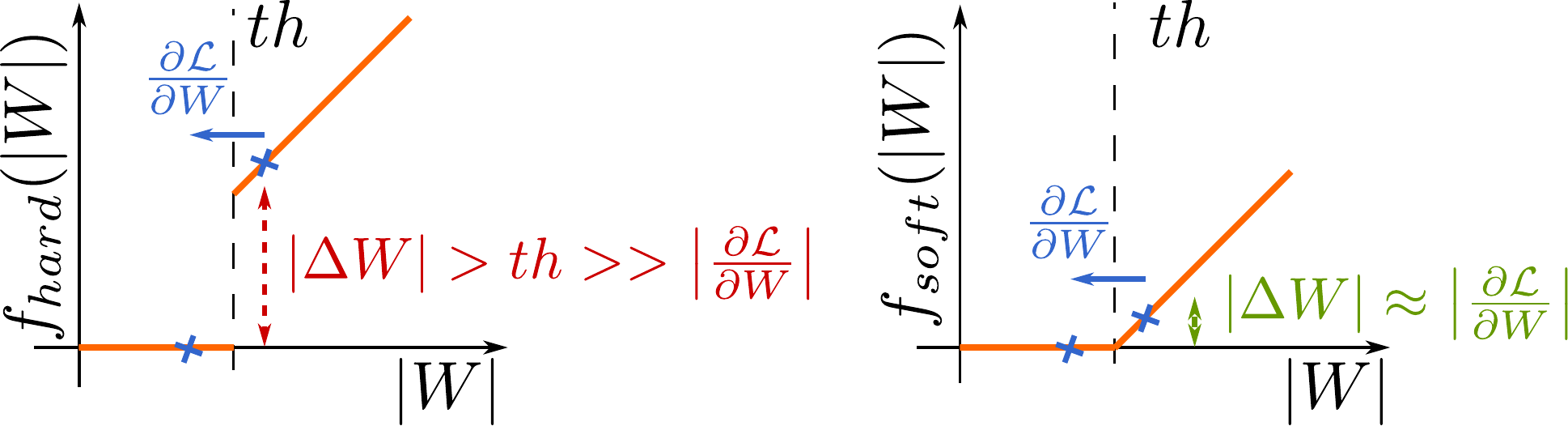}
   \caption{Hard-thresholding (left) induces a discrepancy between the forward weight update and the gradient received in the backward path. Soft-thresholding (right) does not suffer from such inconsistency, since it preserves a smooth relationship between update and gradient value. This prevents sharp weight discontinuities along training.}
   \label{fig:soft}
\end{figure}
Most of the persistent pruning literature makes use of hard thresholding to increase the pruning ratio \cite{Han2015,Frankle2019,Evci2020}. However, in sparse training, since the weights that are zeroed (because they are below the threshold) at some point during the training might become significant again in subsequent training iterations, they are also likely to cross the pruning threshold, and be subject to the hard thresholding function discontinuity. This might lead to abrupt changes in the forward path, which might be inconsistent with the gradient received in the backward path. Figure \ref{fig:soft} depicts a case where the pruning threshold is large compared to the gradient magnitude.
In this case, hard thresholding induces severe discontinuities in the forward weight update, while soft thresholding preserves a smooth and consistent evolution of this forward weight. The results shown in Figure \ref{fig:ste} confirm that soft thresholding leads to a more stable training, and better accuracy of trained models.

\begin{figure*}
  \centering
\includegraphics[width=0.49\linewidth]{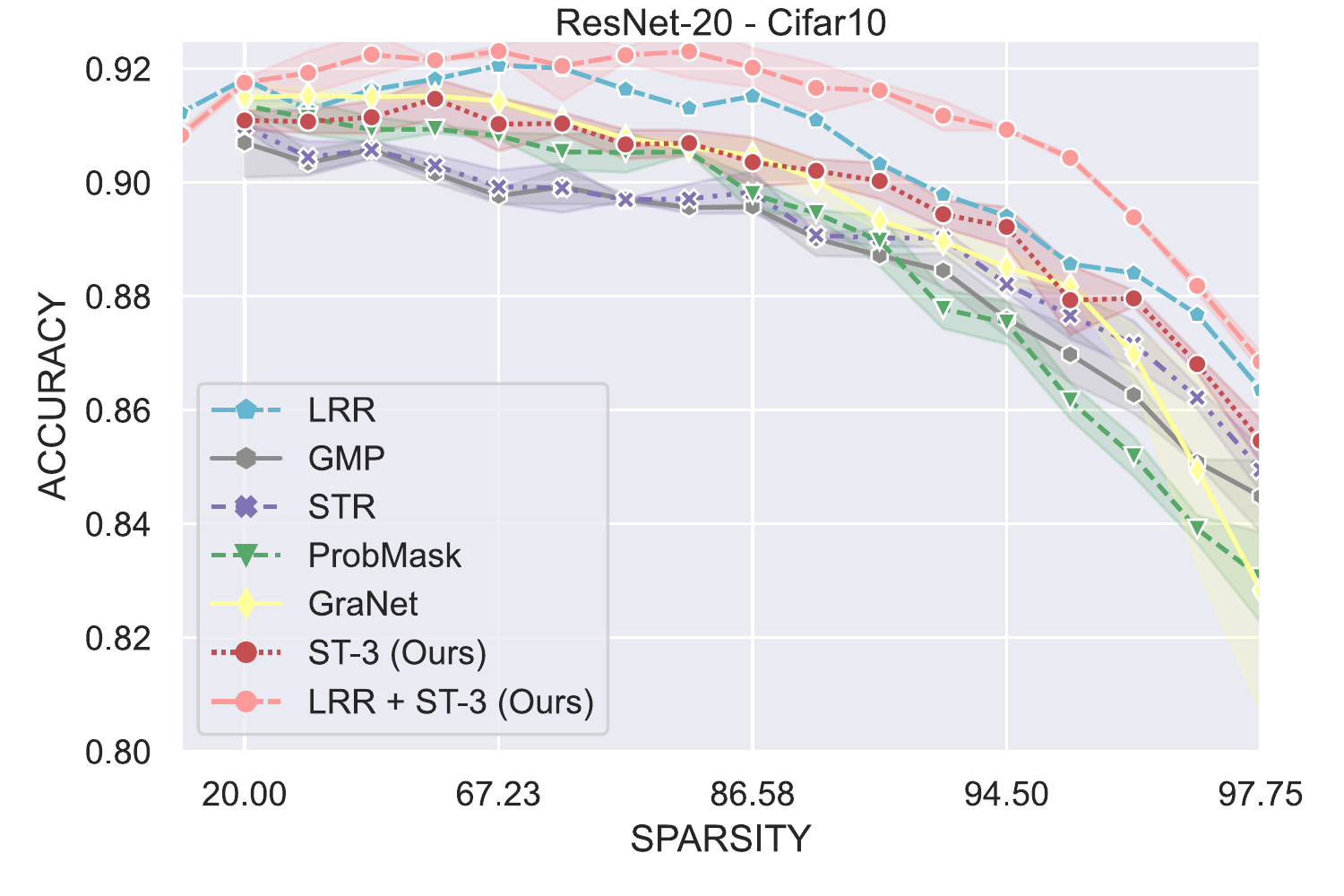}
\includegraphics[width=0.49\linewidth]{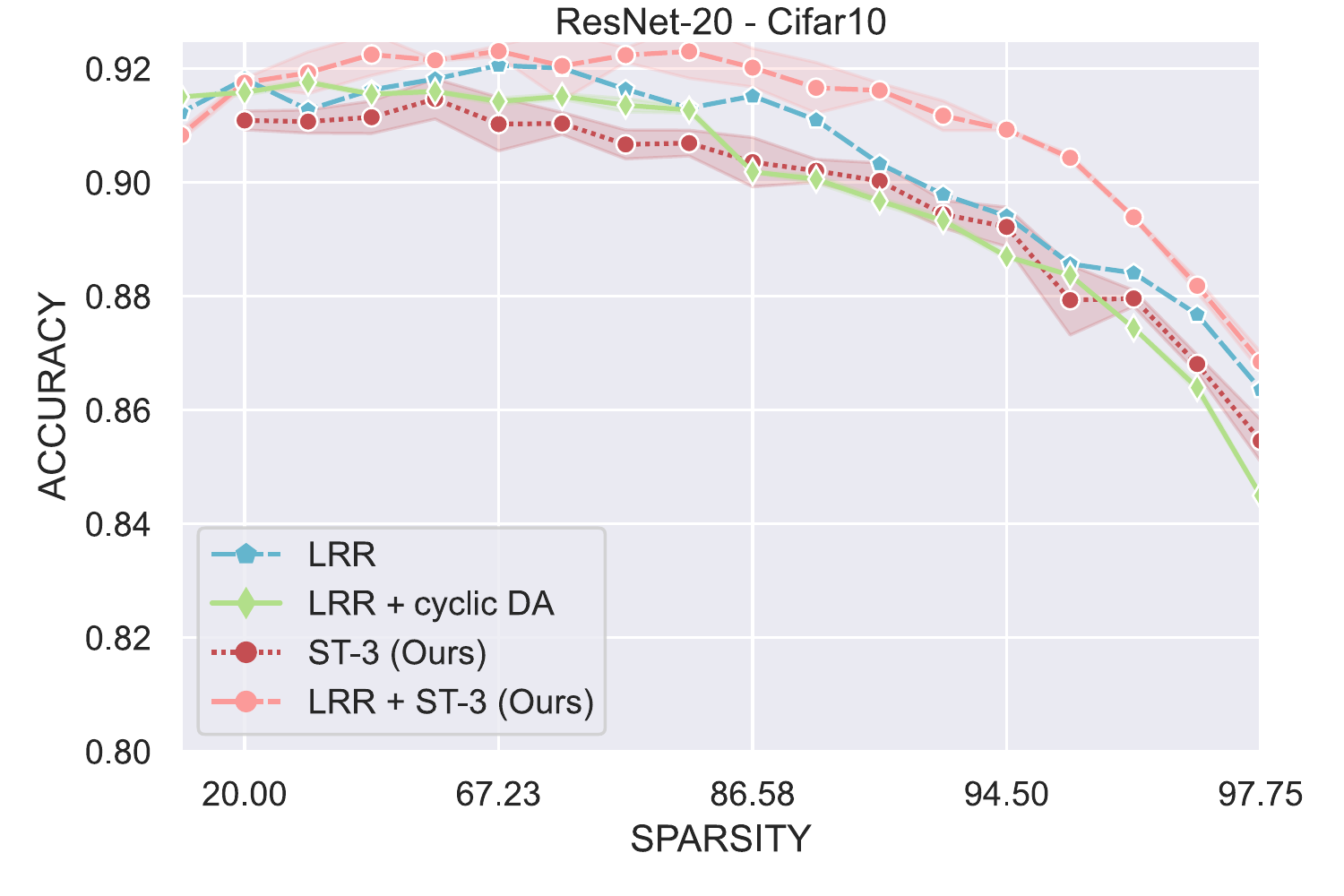}\\

\includegraphics[width=0.49\linewidth]{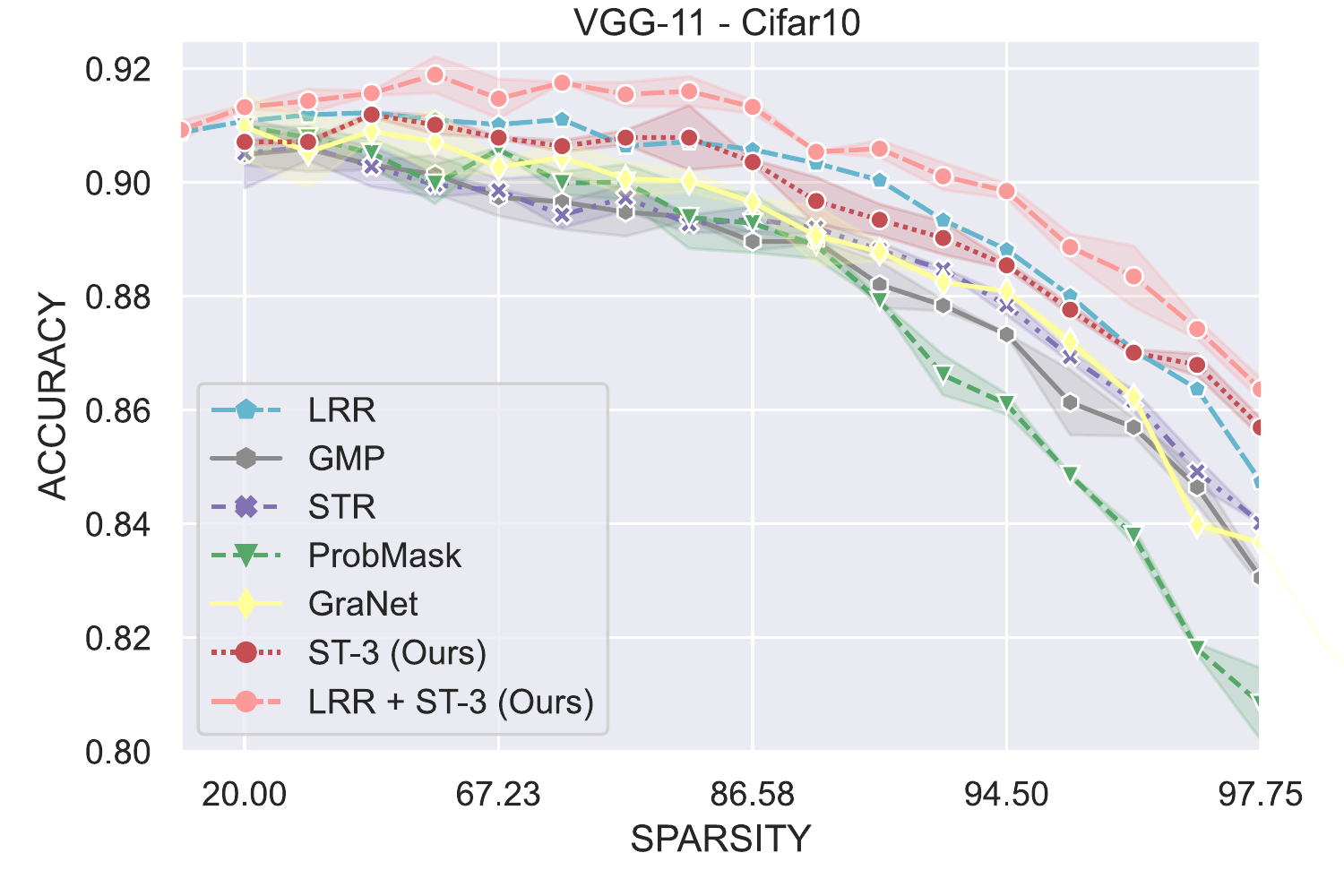}
\includegraphics[width=0.49\linewidth]{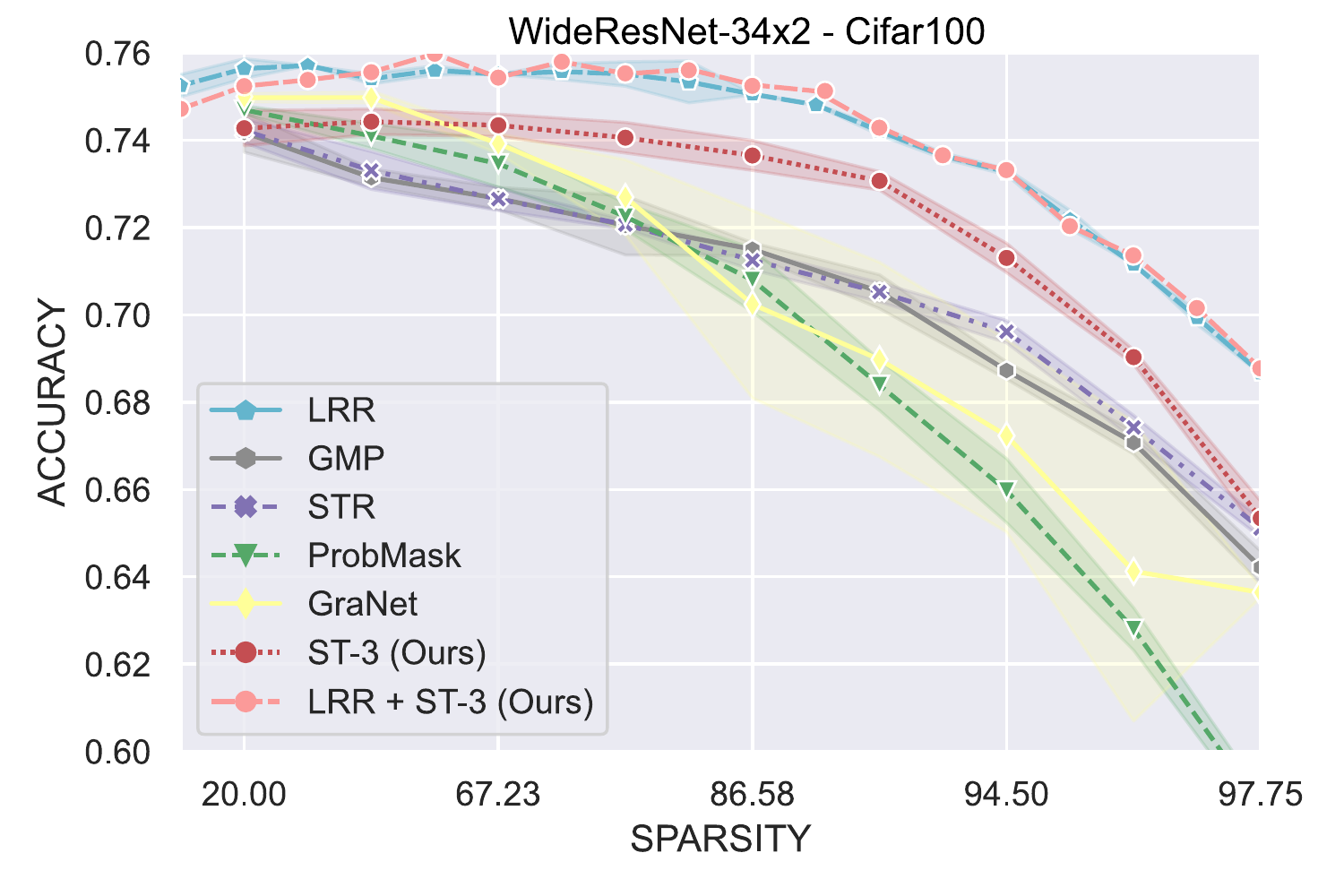}

  \caption{\textit{(best viewed in color)} (top) Accuracy/sparsity curves obtained, on Cifar10 w/ ResNet-20, (bottom) Accuracy/sparsity curves obtained on Cifar10 w/ VGG-11, and on Cifar100 w/ WideResNet-34x2. The methods starting w/ LRR are recursively trained, i.e. each point on the curve needs every point on its left to be computed first. All points represent the average of 3 runs (with different seeds common to all methods) and standard deviation is shown. The x-axis is depicted in log-scale.
  }
  \label{fig:cifar}
\end{figure*}

\paragraph{Weight rescaling.}
The weights that contribute to the same output neuron are referred to as being part of the same \emph{filter}. 
To preserve the mean absolute magnitude of a neuron in presence of weight zeroing, we propose to rescale the active weights filter-wise, based on the ratio of lost weight-magnitude introduced by the zeroed weights. Formally, letting $W^l \in R^{Nout \times Nin}$ denote the dense weight matrix (before soft-thresholding) of layer $l$ and $th$ be the sparsity threshold, then the factor multiplying the soft-thresholded vector associated to the $j^{th}$ output is defined as follows:

\begin{footnotesize}
$$
\text{scale}^l_j = \displaystyle\frac{\sum\limits_i^{Nin}~\|W^l_{ji}\|}
{\sum\limits^{Nin}_i~\begin{cases}
\|W^l_{ji}\| & \quad \text{if }  \|W^l_{ji}\| > th\\
0 & \quad \text{otherwise}
\end{cases}}
$$
\end{footnotesize}


\noindent This is similar to the rescaling used by dropout \cite{srivastava14a}. 

\paragraph{Increase and layer-wise allocation of sparsity.}
To mitigate the instabilities induced by the thresholding, and leave the network the time to adapt to the deactivation of some weights in the forward path, we increase the sparsity ratio progressively along training. 
To control the increase of global sparsity, the cubic pattern introduced in \cite{Zhu2018} is adopted. Two variants of our method have been studied in our experiments to turn the global sparsity ratio into a layerwise sparsity.\\
The first variant, denoted {\bf ST-3}, is adopted by default. It simply considers the network weights globally. Hence, no specific attention is drawn to the balance of sparsity across layers. It has the advantage of being straightforward, and our experiments reveal that it results in better accuracy/sparsity tradeoffs as ERK or LAMP.\\
The second variant, denoted {\bf ST-3$^\sigma$}, has been designed to bias the training towards increased FLOPS gains.
It arises from the observation that to save FLOPS, it is better to prune weights in layers where the feature maps are larger, since large maps require more operations. However, 
large feature maps generally manipulate small kernels (i.e. due to the adoption of a small number of high-resolution channels in common architectures), whose initialization results in large weight magnitudes when adopting the popular Kaiming He initialization \cite{He2015a}. As a consequence, as shown experimentally \cite{Evci2020}, after global thresholding, those kernels with larger weights end-up in being relatively less sparse than the kernels associated to the smaller resolutions channels of wider layers.
This aspect is generally overlooked in previous work \cite{Frankle2020,Frankle2019,Renda2020,Lee2021}, where only the model compression is considered, disregarding the number of operations affected by the zeroed weights. Hence, most works \cite{Kusupati2020,Zhou2021} treat inference acceleration as a by-product, while still optimizing only for a specified sparsity ratio. To favor the zeroing of weights in layers manipulating (few) high resolution channels, and improve the computational gain associated to sparsity, we propose, before soft-thresholding, to multiply each weight by the square root of the number of weights in its corresponding kernel. This can be interpreted as normalizing the weights by their standard deviation in individual kernel since the variance of the weights following the Kaiming He initialization distribution is known to be inversely proportional to the size of the kernel \cite{He2015a}.  

\section{Experiments}\label{sec:experiments}
\subsection{Methods used as baselines}\label{sec:methods}
Our approach is compared to a set of previous methods that are representative of the SoA and of the recent trends in the field. We now introduce those methods, and explain how they have been configured to provide a fair comparison, and help in quantifying the actual benefit brought by our proposed ST-3.

As a recognized SoA upper bound in terms of accuracy/sparsity trade-off, the Learning-Rate-Rewind (LRR) recursive pruning method is considered \cite{Renda2020}. This method however has the obvious downside of requiring significantly more training cycles than every other method considered in this paper. Typically 10 consecutive training cycles are needed to reach 90\% sparsity, which isn't practically scalable to larger datasets. In our experiments, LRR is also combined with our ST-3 as follows: the cubic increase of sparsity ratio adopted in ST-3 is replaced by the increase of sparsity ratio implemented by LRR between training cycles, with learning rate rewind. This hybrid approach is denoted LRR+ST-3. It is as expensive to train than LRR, but leads by far to the best accuracy/sparsity trade-offs, which confirms the benefit of straight-through estimation combined to soft-thresholding with magnitude loss compensation.

Among the methods adopting a single training cycle, the GMP \cite{Zhu2018}, STR \cite{Kusupati2020}, and ProbMask \cite{Zhou2021} methods, presented in Section 2, are probably the most relevant to compare with: GMP and STR because they are conceptually simple threshold-based pruning and sparsification methods, respectively; ProbMask because it corresponds to the current SoA in one-training cycle methods (even if its training time is twice as long as the ST-3 one, due to gradient accumulation required by the Gumbel-Softmax trick at each training step).

For both GMP and STR methods, the \emph{sparsity ratio} increases following the same cubic polynomial function as the one adopted by our method (see Sec. \ref{sec:ours}), and global thresholding is adopted to achieve the required sparsity ratio, as for ST-3 \footnote{Homogenizing the methods configuration does not significantly impact the method performance (as demonstrated by our results on ImageNet in Sec. \ref{sec:imagenet})}.
For ProbMask, the sparsity increase ratio and batch-sizes are kept as in the original paper, as changing them led to significant degradation in accuracy. 

For completeness, comparison is also provided with DNW \cite{Wortsman2019}, RIGL \cite{Evci2020} and GraNet \cite{Liu2021} (see Section 2 for the presentation of those methods), implemented as recommended by their respective authors.

\begin{figure*}
  \centering
    \includegraphics[width=0.49\linewidth]{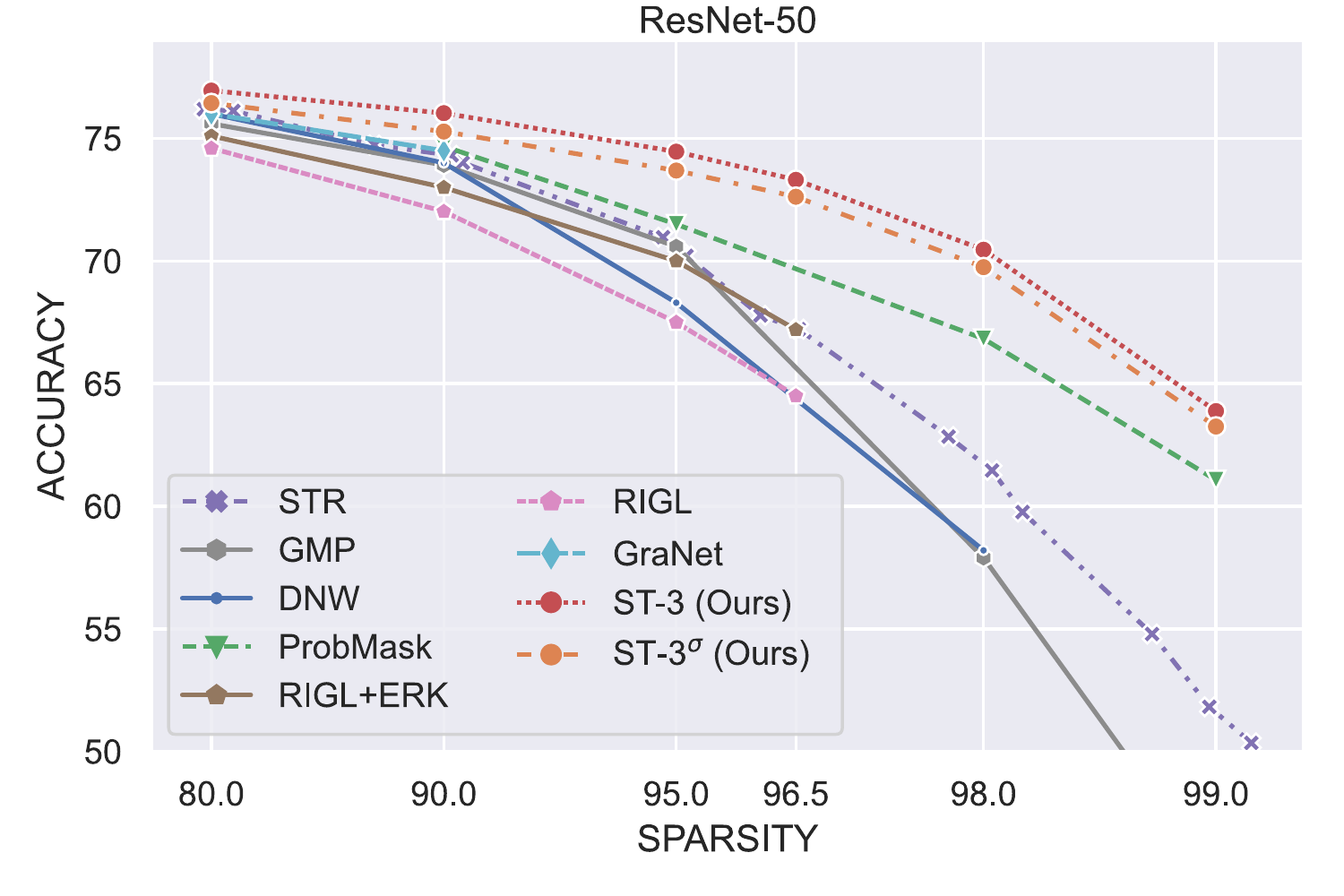}
    \includegraphics[width=0.49\linewidth]{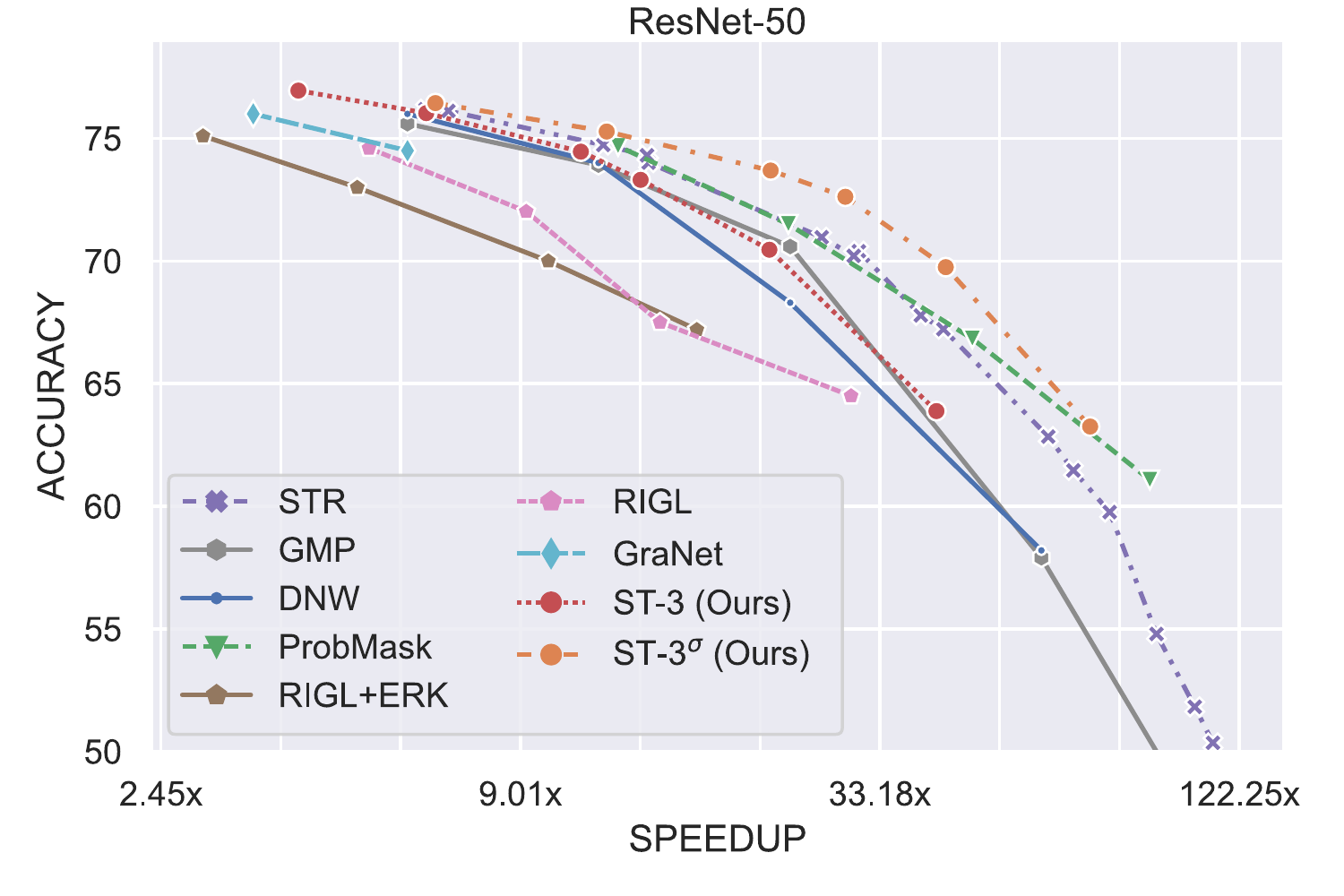}\\
    \includegraphics[width=0.47\linewidth]{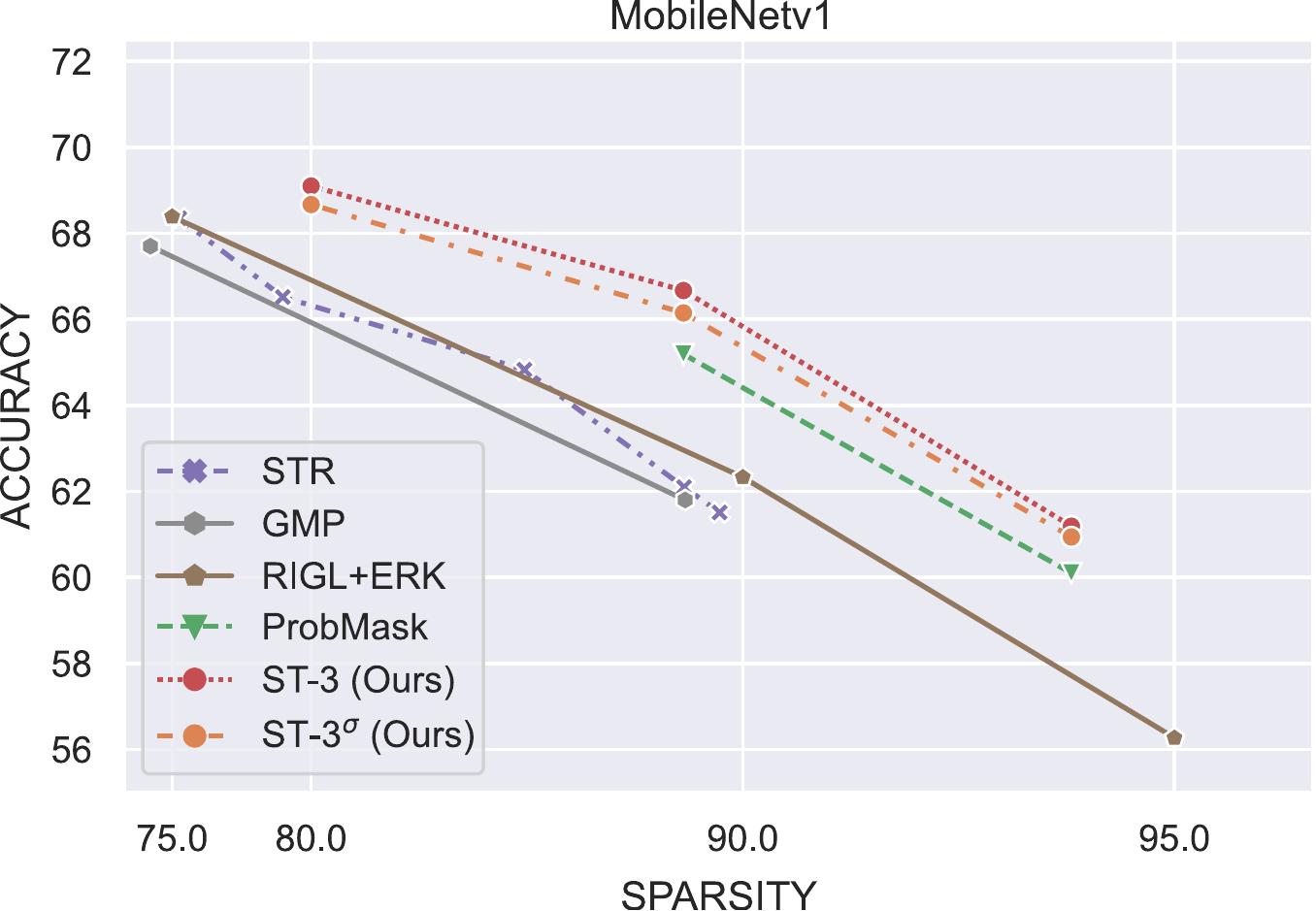}~~
    \includegraphics[width=0.47\linewidth]{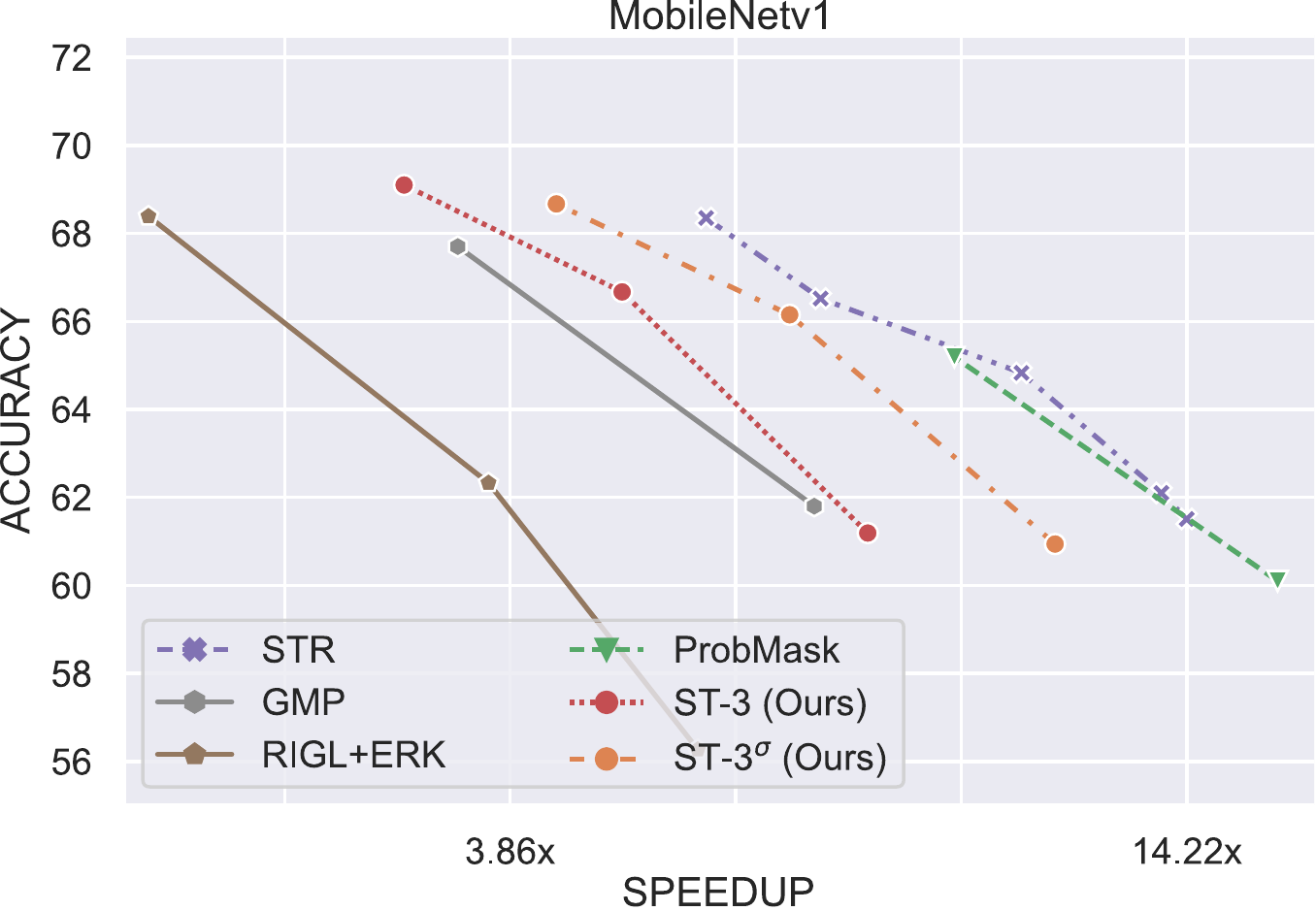}
  \caption{Accuracy on ImageNet w/ ResNet-50 (top) and MobileNetv1 (bottom) as a function of the global sparsity (left), and of the FLOPS (right). All graphs use a log-scale for the x-axis.}\label{fig:imagenet}
\end{figure*}

\label{sec:cifar}

\subsection{Cifar10 and Cifar100}

The methods introduced in Section \ref{sec:methods} have been run on Cifar-10 and Cifar-100 \cite{Krizhevsky2009}. Three different networks are considered: ResNet-20 \cite{He2015}, VGG-11 \cite{Simonyan2014}) (as simplified by \cite{Liu2019}), and WideResNet-34x2 \cite{ZagoruykoK16}.
Similar to what is done in related papers, one training cycle consists in 160 epochs and, in the single training cycle scenario, the progressive increase of the sparsity ratio along a training cycle starts at the 5$^{th}$ epoch and ends at the 80$^{th}$ epoch.

\paragraph{One training cycle.} In Figure \ref{fig:cifar}, our ST-3 achieves the best accuracy/sparsity tradeoffs among methods trained with a single training cycle, including compared to ProbMask\cite{Zhou2021}\footnote{Note that, in contrast to what is implemented in the code released with the original paper, our results have been generated by adapting the network's batch-normalization moving-mean and moving-variance on the training – and not the testing - set. Adapting on the testing set indeed raises a methodological issue. This difference explains the slight discrepancy between our plots and the ones in \cite{Zhou2021}.} and GraNet\cite{Liu2021}. The soft-thresholding mechanism adopted in STR \cite{Kusupati2020} makes it possible for some weights to move between inactive and active states, which results in a significant boost in performance compared to GMP \cite{Zhu2018}. However, STR falls short compared to ST-3. This is because STR does not account for gradient information in any way, except for momentum-like terms stored in the optimizer. ST-3 solves this issue, by allowing the accumulation of the gradient information conveyed by the straight-through-estimator.  

\paragraph{Multiple training cycles.} The computational cost inherent to the multiple training cycles imposed by LRR makes it impractical. However, LRR remains a de-facto upper bound to compare with when considering the training of sparse network. Without surprise, we observe in Figure \ref{fig:cifar} that LRR performs better than ST-3. More interestingly, LRR+ST-3 method is even better, defining a new SoA. The superiority of LRR+ST-3 over LRR provides an additional clue that the use of a straight-through estimator to continuously and smoothly update the raw (non-thresholded) version of zeroed weights leads to a significant increase in performance when training sparse networks. A deeper comparison between ST-3 and LRR reveals that ST-3 gets closer to LRR on Cifar10 than with Cifar100. We attribute this phenomenon to the fact that the increased data augmentation associated to the longer training implemented  by LRR is more beneficial for Cifar100 (only 500 training images per class) than for Cifar10 (5000 training images per class). To challenge this hypothesis, we have trained ST-3 for 4x more epochs than a reference cycle at a sparsity threshold of 67.23\% on ResNet-20. This longer training has resulted in a test accuracy of 92.38\% which is better than the 92.22\% obtained by LRR with 6x the number of epochs of a reference training cycle.  
The role of data augmentation in the performance improvement observed with longer training is also explored on ResNet-20 in Figure \ref{fig:cifar} \textit{(top right)}. LRR is applied as described earlier, but the augmentation seed is reset between each training cycle, ensuring that LRR manipulates the same augmented samples than one-cycle methods. 
With this additional constraint, we observe that the LRR performance degrades up to the one of our ST-3.

\subsection{ImageNet}\label{sec:imagenet}
ResNet-50 \cite{He2015} and MobileNetv1 \cite{howard2017mobilenets} have been trained on ImageNet \cite{deng2009imagenet} using the same standard hyper-parameters, number of epochs (=100), and data-augmentation as previous related works \cite{Evci2020,Kusupati2020,Zhou2021,Liu2021}, leading to a \emph{dense} baseline accuracy of 77.1\%. To train sparse models, only one-cycle methods have been considered, due to the excessive training time required by the LRR iterations. The sparsity progressively increases between the 5$^{th}$ and the 50$^{th}$ epochs. Values for curves corresponding to related works with ResNet-50 on ImageNet are directly copied from their respective papers \footnote{In particular, this means that the ProbMask accuracy is the one obtained in \cite{Zhou2021}, i.e. when using test samples to define the batch-normalization parameters. Hence, they correspond to an upper bound of the results that would be obtained when defining batch-normalization parameters with training samples only. 
}. 

\paragraph{Accuracy/Sparsity}
In Figure \ref{fig:imagenet}, ST-3 reaches the best accuracy/sparsity tradeoffs. The closest contenders are the most recent results provided by GraNet and ProbMask, but a consistent gap remains that widens up to 5\% at 95\% sparsity. Our methods unequivocally pull ahead when optimising sparsity only. 
It is interesting to note that the gap between the two variants of our method, namely ST-3 and ST-3$^\sigma$, is moderate and stays more or less constant across all sparsity ratios. The superiority of ST-3 over ST-3$^\sigma$ confirms previous observations indicating that pruning earlier layers penalizes accuracy (slightly) more than zeroing later ones \cite{Evci2020}. 

\paragraph{Accuracy/FLOPS}
FLOPS is a widespread metric to assess the computational gain corresponding to an ideal scenario where every multiplication with zero can be avoided. In practice, inference speedup is measured as the FLOPS reduction factor. 
As depicted in Figure \ref{fig:imagenet}, our ST-3$^\sigma$ method, a ST-3 variant proposed to bias the distribution of sparsity across layers in favor of layers with higher resolution, significantly improves sparsity/speedup tradeoffs, making our approach competitive in that respect as well.

\subsection{Ablation studies}\label{sec:ablation}
\begin{figure}
  \centering
    \includegraphics[width=0.98\linewidth]{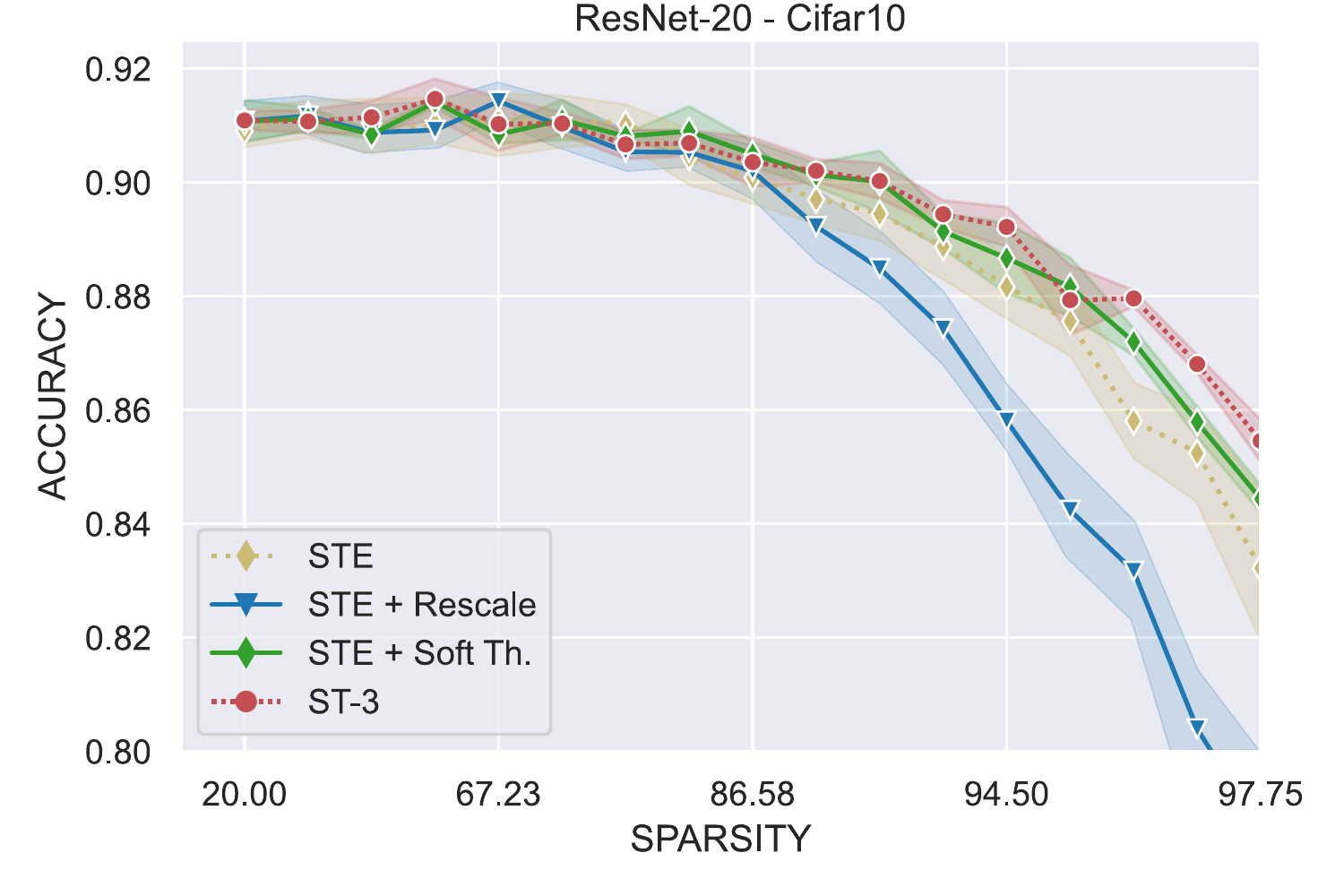}
  \caption{A study dissecting the impact of straight-through estimation, soft-thresholding and weight-rescaling during sparse training. A log-scale is used for the x-axis.}\label{fig:ablation}
\end{figure}
\paragraph{Isolating ST-3 components} Figure \ref{fig:ablation} evaluates the importance of the three components of our method, namely straight-through-estimation (STE), soft-thresholding, and weight rescaling for mean magnitude loss compensation. Straight-through-estimation and soft-thresholding appear to be the main contributors to the success of our method. The rescaling factor helps in some respect but only when soft-thresholding is used. 
When used with pure STE, meaning with hard-thresholding, it tends to amplify the number of sharp discontinuities in the evolution of weights, thereby causing a severe penalty.

\paragraph{Weight switching between active/inactive states}
\begin{figure*}[t]
    
  \centering
    \includegraphics[width=0.49\linewidth]{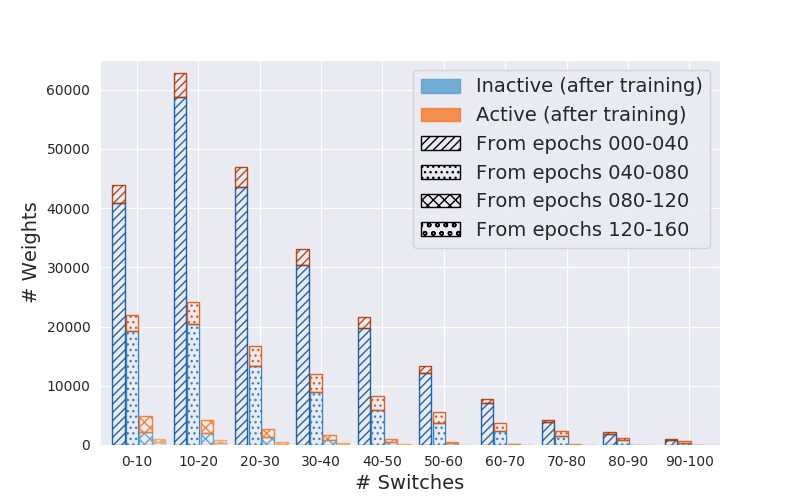}
    \includegraphics[width=0.49\linewidth]{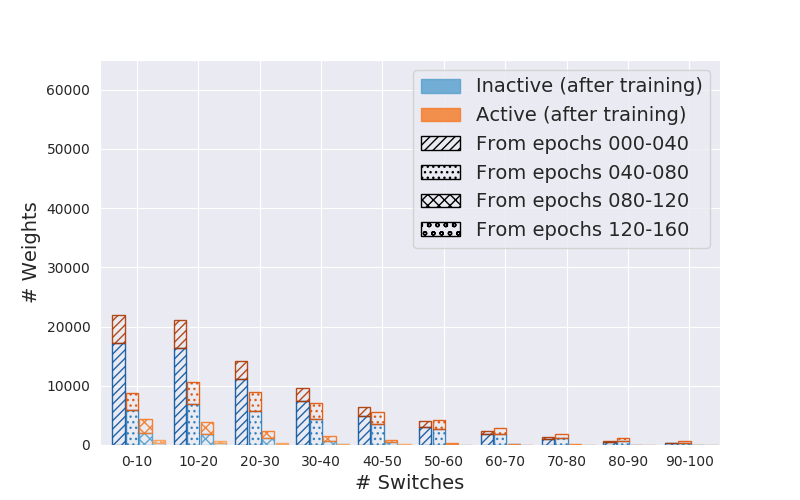}
  \caption{\textit{(best viewed in color)} Number of switches between active and inactive state for the weights of a ResNet-20 trained with ST-3 for 160 epochs on Cifar-10 at 90\% sparsity. The global sparsity ratio is fixed (left) or cubicly increasing (right). (Weights that switch once or never are not included)}
  \label{fig:thcrossings}
\end{figure*}

Figure \ref{fig:thcrossings} investigates how the soft-thresholded weights in ST-3 \emph{switch} between zero and non-zero values during training. It presents how the counts of switches per weight are distributed in four sequences of epochs, obtained by partitioning the total number of epochs in groups of 40 epochs.  For clarity, weights that switch less than two times are ignored in the plot. When plotting these distributions, a color code is used to distinguish the weights that  are active (i.e. non-zero) at the end of training from the ones that are zeroed.

On the left, the sparsity ratio is set to its final value from the beginning of training, while it progressively increases along training on the right.
We observe that progressively increasing the sparsity ratio fundamentally changes the weight switching pattern. Progressive increase stabilize the training, as attested by the reduced amount of switches between states, especially in the second group of epochs. It also makes the state switching pattern closer to the typical sign switching pattern observed in dense training (the signs of weights that remain significant at the end of training change more often in the first epochs). A progressive increase of the sparsity is thus desired since it complements the stability offered by straight-through estimation and soft-thresholding. \\

\section{Conclusion}
Throughout this paper the relevance of our straightforward and computationally simple approach, ST-3$^{(\sigma)}$, has been widely demonstrated. Its efficiency in training very sparse networks has been assessed on both cifar-10(0) and ImageNet with a variety of architectures, attaining single-cycle SoA accuracy. 
The fact that our simple and relatively straightforward approach achieves the highest accuracies among approaches with comparable complexity suggests that the key towards effective sparsification does not hold in a sophisticated formulation of the optimization problem but rather in the ability to give the weights the freedom to evolve across the zero state while progressively increasing the sparsity ratio.

{\small
\bibliographystyle{ieee_fullname}
\bibliography{main}

\begin{thebibliography}{10}\itemsep=-1pt

\bibitem{Bengio13}
Yoshua Bengio, Nicholas L{\'{e}}onard, and Aaron~C. Courville.
\newblock Estimating or propagating gradients through stochastic neurons for
  conditional computation.
\newblock {\em CoRR}, abs/1308.3432, 2013.

\bibitem{deng2009imagenet}
Jia Deng, Wei Dong, Richard Socher, Li-Jia Li, Kai Li, and Li Fei-Fei.
\newblock Imagenet: A large-scale hierarchical image database.
\newblock In {\em 2009 IEEE conference on computer vision and pattern
  recognition}, pages 248--255. Ieee, 2009.

\bibitem{Evci2020}
Utku Evci, Trevor Gale, Jacob Menick, Pablo~Samuel Castro, and Erich Elsen.
\newblock {Rigging the lottery: Making all tickets winners}.
\newblock {\em 37th International Conference on Machine Learning, ICML 2020},
  PartF16814:2923--2933, 2020.

\bibitem{Frankle2019}
Jonathan Frankle and Michael Carbin.
\newblock {The lottery ticket hypothesis: Finding sparse, trainable neural
  networks}.
\newblock {\em 7th International Conference on Learning Representations, ICLR
  2019}, pages 1--42, 2019.

\bibitem{Frankle2020b}
Jonathan Frankle, Gintare~Karolina Dziugaite, Daniel Roy, and Michael Carbin.
\newblock Pruning neural networks at initialization: Why are we missing the
  mark?
\newblock {\em ICLR}, 2021.

\bibitem{Frankle2020}
Jonathan Frankle, David~J Schwab, and Ari~S Morcos.
\newblock {The Early Phase of Neural Network Training}.
\newblock In {\em International Conference on Learning Representations}, pages
  1--20, 2020.

\bibitem{Gale2019}
Trevor Gale, Erich Elsen, and Sara Hooker.
\newblock The state of sparsity in deep neural networks.
\newblock {\em CoRR}, abs/1902.09574, 2019.

\bibitem{Gray2017}
Scott Gray, Alec Radford, and Diederik~P Kingma.
\newblock {GPU Kernels for Block-Sparse Weights}.
\newblock {\em OpenAI '17}, (1):12.

\bibitem{Han2015}
Song Han, Jeff Pool, John Tran, and William~J. Dally.
\newblock {Learning both weights and connections for efficient neural
  networks}.
\newblock {\em Advances in Neural Information Processing Systems},
  2015-Janua:1135--1143, 2015.

\bibitem{He2015}
Kaiming He, Xiangyu Zhang, Shaoqing Ren, and Jian Sun.
\newblock {Deep Residual Learning for Image Recognition}.
\newblock 37(50):1951--1954, dec 2015.

\bibitem{He2015a}
Kaiming He, Xiangyu Zhang, Shaoqing Ren, and Jian Sun.
\newblock Delving deep into rectifiers: Surpassing human-level performance on
  imagenet classification.
\newblock {\em CoRR}, abs/1502.01852, 2015.

\bibitem{He2019a}
Yang He, Ping Liu, Ziwei Wang, Zhilan Hu, and Yi Yang.
\newblock {Filter pruning via geometric median for deep convolutional neural
  networks acceleration}.
\newblock {\em Proceedings of the IEEE Computer Society Conference on Computer
  Vision and Pattern Recognition}, 2019-June:4335--4344, 2019.

\bibitem{howard2017mobilenets}
Andrew~G Howard, Menglong Zhu, Bo Chen, Dmitry Kalenichenko, Weijun Wang,
  Tobias Weyand, Marco Andreetto, and Hartwig Adam.
\newblock Mobilenets: Efficient convolutional neural networks for mobile vision
  applications.
\newblock {\em arXiv preprint arXiv:1704.04861}, 2017.

\bibitem{Hubara2016}
Itay Hubara, Matthieu Courbariaux, Daniel Soudry, Ran El-Yaniv, and Yoshua
  Bengio.
\newblock {Quantized Neural Networks: Training Neural Networks with Low
  Precision Weights and Activations}.
\newblock 2016.

\bibitem{Kalchbrenner2018}
Nal Kalchbrenner, Erich Elsen, Karen Simonyan, Seb Noury, Norman Casagrande,
  Edward Lockhart, Florian Stimber, A{\"{a}}ron {Van Den Oord}, Sander
  Dieleman, and Koray Kavukcuoglu.
\newblock {Efficient neural audio synthesis}.
\newblock {\em 35th International Conference on Machine Learning, ICML 2018},
  6:3775--3784, 2018.

\bibitem{Krizhevsky2009}
Alex Krizhevsky.
\newblock {(CIFAR10) Learning Multiple Layers of Features from Tiny Images}.
\newblock {\em {\ldots} Science Department, University of Toronto, Tech.
  {\ldots}}, pages 1--60.

\bibitem{Kusupati2020}
Aditya Kusupati, Vivek Ramanujan, Raghav Somani, Mitchell Wortsman, Prateek
  Jain, Sham Kakade, and Ali Farhadi.
\newblock Soft threshold weight reparameterization for learnable sparsity.
\newblock pages 5544--5555, 2020.

\bibitem{Leclerc2020}
Guillaume Leclerc and Aleksander Madry.
\newblock {The Two Regimes of Deep Network Training}.
\newblock 2020.

\bibitem{LeCun1990}
Yann LeCun, John~S Denker, and Sara~A. Solla.
\newblock {Optimal Brain Damage (Pruning)}.
\newblock {\em Advances in neural information processing systems}, pages
  598--605, 1990.

\bibitem{Lee2021}
Jaeho Lee, Sejun Park, Sangwoo Mo, Sungsoo Ahn, and Jinwoo Shin.
\newblock Layer sparsity for the magnitude-based pruning.
\newblock {\em CoRR}, abs/2010.07611, 2020.

\bibitem{Lee2019}
Namhoon Lee, Thalaiyasingam Ajanthan, and Philip~H.S. Torr.
\newblock {SnIP: Single-shot network pruning based on connection sensitivity}.
\newblock In {\em 7th International Conference on Learning Representations,
  ICLR 2019}, 2019.

\bibitem{Li2020}
Bailin Li, Bowen Wu, Jiang Su, and Guangrun Wang.
\newblock {EagleEye: Fast Sub-net Evaluation for Efficient Neural Network
  Pruning}.
\newblock {\em Lecture Notes in Computer Science (including subseries Lecture
  Notes in Artificial Intelligence and Lecture Notes in Bioinformatics)}, 12347
  LNCS:639--654, 2020.

\bibitem{Li2016}
Fengfu Li, Bo Zhang, and Bin Liu.
\newblock {Ternary Weight Networks}.
\newblock (Nips), 2016.

\bibitem{Li2017}
Hao Li, Hanan Samet, Asim Kadav, Igor Durdanovic, and Hans~Peter Graf.
\newblock {Pruning filters for efficient convnets}.
\newblock {\em 5th International Conference on Learning Representations, ICLR
  2017 - Conference Track Proceedings}, (2016):1--13, 2017.

\bibitem{Liu2021}
Shiwei Liu, Tianlong Chen, Xiaohan Chen, Zahra Atashgahi, Lu Yin, Huanyu Kou,
  Li Shen, Mykola Pechenizkiy, Zhangyang Wang, and Decebal~Constantin Mocanu.
\newblock {Sparse Training via Boosting Pruning Plasticity with
  Neuroregeneration}.
\newblock (NeurIPS), 2021.

\bibitem{liu2021we}
Shiwei Liu, Lu Yin, Decebal~Constantin Mocanu, and Mykola Pechenizkiy.
\newblock Do we actually need dense over-parameterization? in-time
  over-parameterization in sparse training.
\newblock In {\em International Conference on Machine Learning}, pages
  6989--7000. PMLR, 2021.

\bibitem{Liu2019}
Zhuang Liu, Mingjie Sun, Tinghui Zhou, Gao Huang, and Trevor Darrell.
\newblock {Rethinking the value of network pruning}.
\newblock {\em 7th International Conference on Learning Representations, ICLR
  2019}, pages 1--21, 2019.

\bibitem{Mellempudi2017}
Naveen Mellempudi, Abhisek Kundu, Dheevatsa Mudigere, Dipankar Das, Bharat
  Kaul, and Pradeep Dubey.
\newblock {Ternary Neural Networks with Fine-Grained Quantization}.
\newblock 2017.

\bibitem{Mishra2021}
Asit~K. Mishra, Jorge~Albericio Latorre, Jeff Pool, Darko Stosic, Dusan Stosic,
  Ganesh Venkatesh, Chong Yu, and Paulius Micikevicius.
\newblock Accelerating sparse deep neural networks.
\newblock {\em CoRR}, abs/2104.08378, 2021.

\bibitem{Mocanu2018}
Decebal~Constantin Mocanu, Elena Mocanu, Peter Stone, Phuong~H. Nguyen,
  Madeleine Gibescu, and Antonio Liotta.
\newblock {Scalable training of artificial neural networks with adaptive sparse
  connectivity inspired by network science}.
\newblock {\em Nature Communications}, 9(1):1--12, 2018.

\bibitem{Park2017}
Jongsoo Park, Hai Li, Sheng Li, Wei Wen, Yiran Chen, Ping Tak~Peter Tang, and
  Pradeep Dubey.
\newblock {Faster cnns with direct sparse convolutions and guided pruning}.
\newblock {\em 5th International Conference on Learning Representations, ICLR
  2017 - Conference Track Proceedings}, 2017:1--12, 2017.

\bibitem{Renda2020}
Alex Renda, Jonathan Frankle, and Michael Carbin.
\newblock {Comparing Rewinding and Fine-tuning in Neural Network Pruning}.
\newblock (2019):1--31, 2020.

\bibitem{Sanh2020}
Victor Sanh, Thomas Wolf, and Alexander~M. Rush.
\newblock {Movement pruning: Adaptive sparsity by fine-tuning}.
\newblock {\em Advances in Neural Information Processing Systems},
  2020-Decem(NeurIPS):1--12, 2020.

\bibitem{Simonyan2014}
K. Simonyan and A. Zisserman.
\newblock {Very Deep Convolutional Networks for Large-Scale Image Recognition}.
\newblock {\em CoRR}, 2014.

\bibitem{srivastava14a}
Nitish Srivastava, Geoffrey Hinton, Alex Krizhevsky, Ilya Sutskever, and Ruslan
  Salakhutdinov.
\newblock Dropout: A simple way to prevent neural networks from overfitting.
\newblock {\em Journal of Machine Learning Research}, 15(56):1929--1958, 2014.

\bibitem{Tanaka2020}
Hidenori Tanaka, Daniel Kunin, Daniel~L.K. Yamins, and Surya Ganguli.
\newblock {Pruning neural networks without any data by iteratively conserving
  synaptic flow}.
\newblock {\em Advances in Neural Information Processing Systems},
  2020-Decem(NeurIPS), 2020.

\bibitem{Wang2020}
Huan Wang, Can Qin, Yulun Zhang, and Yun Fu.
\newblock {Neural Pruning via Growing Regularization}.
\newblock pages 1--15, 2020.

\bibitem{Wortsman2019}
Mitchell Wortsman, Ali Farhadi, and Mohammad Rastegari.
\newblock {Discovering neural wirings}.
\newblock {\em Advances in Neural Information Processing Systems}, 32(NeurIPS),
  2019.

\bibitem{Penghang2019}
Penghang Yin, Jiancheng Lyu, Shuai Zhang, Stanley~J. Osher, Yingyong Qi, and
  Jack Xin.
\newblock Understanding straight-through estimator in training activation
  quantized neural nets.
\newblock {\em ICLR}, 2019.

\bibitem{ZagoruykoK16}
Sergey Zagoruyko and Nikos Komodakis.
\newblock Wide residual networks.
\newblock {\em CoRR}, abs/1605.07146, 2016.

\bibitem{Zhou2019a}
Hattie Zhou, Janice Lan, Rosanne Liu, and Jason Yosinski.
\newblock {Deconstructing Lottery Tickets: Zeros, Signs, and the Supermask}.
\newblock (NeurIPS), 2019.

\bibitem{Zhou2021}
Xiao Zhou, Weizhong Zhang, Hang Xu, and Tong Zhang.
\newblock {Effective Sparsification of Neural Networks with Global Sparsity
  Constraint}.
\newblock {\em Proceedings of the IEEE Computer Society Conference on Computer
  Vision and Pattern Recognition}, pages 3598--3607, 2021.

\bibitem{Zhu2018}
Michael~H. Zhu and Suyog Gupta.
\newblock {To prune, or not to prune: Exploring the efficacy of pruning for
  model compression}.
\newblock {\em 6th International Conference on Learning Representations, ICLR
  2018 - Workshop Track Proceedings}, 2018.

\end{thebibliography}
}
\appendix
\section*{Appendix}
\renewcommand{\thesubsection}{\Alph{subsection}}
\thispagestyle{empty}
\begin{figure*}[h!]
  \centering
   \includegraphics[width=0.495\linewidth]{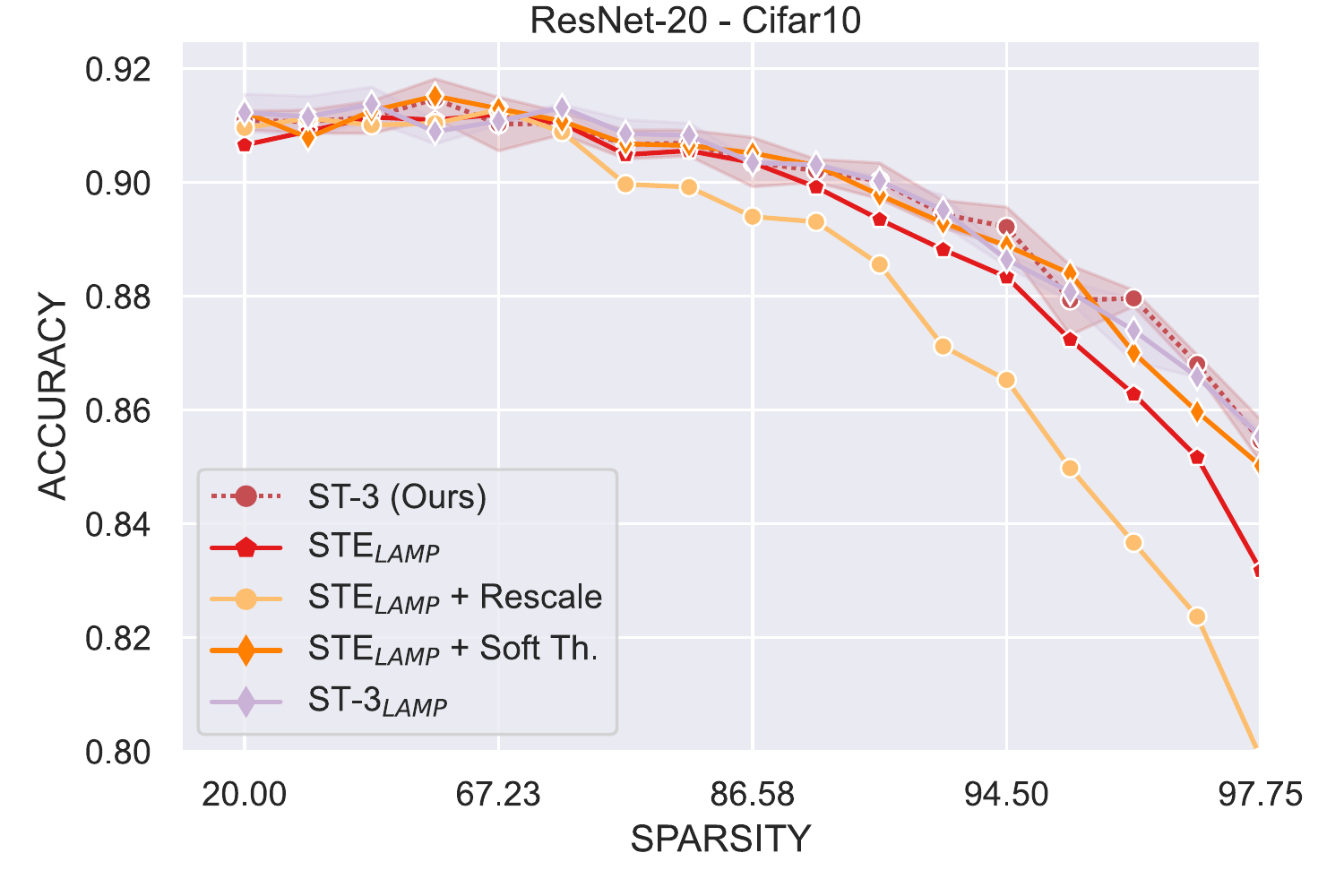}
   \includegraphics[width=0.495\linewidth]{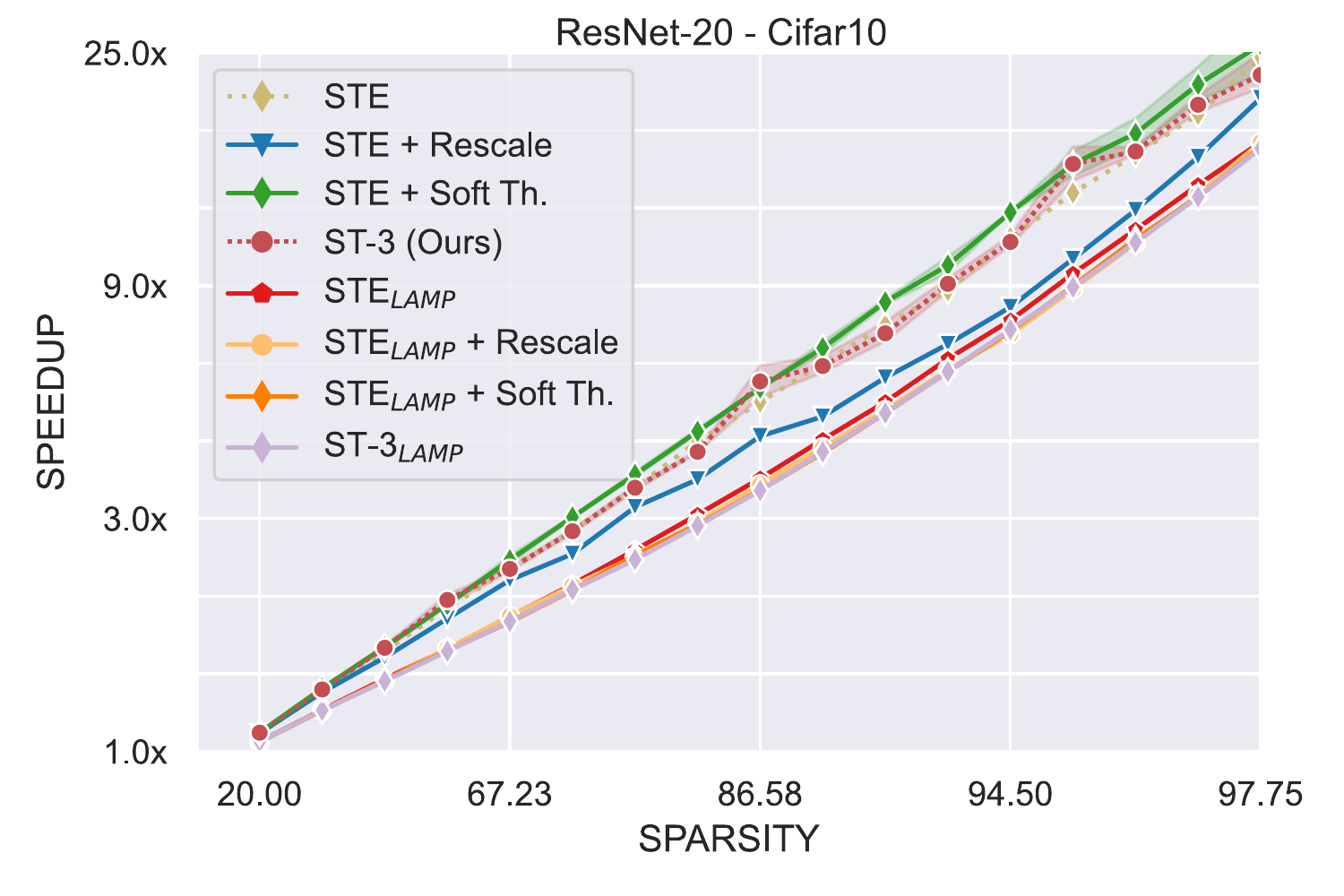}
   \caption{(left) A study dissecting the impact of straight-through estimation, soft-thresholding and weight-rescaling during sparse training with LAMP-based sparse-weight-selection. (right) FLOPS/Sparsity trade-off comparing l1 and LAMP sparse training. A log-scale is used for the x-axes.}
   \label{fig:lamp}
\end{figure*}

This appendix contains the following additional information:
\begin{enumerate}
    \item An additional ablation study on the use of LAMP combined with ST-3
    \item The full hyper-parameter list for our training
    \item Our ST-3$^{(\sigma)}$algorithm in pseudo-code format
    \item The numerical values for the ImageNet results
\end{enumerate}
\subsection{Comparison to LAMP pruning}

Figure \ref{fig:lamp} repeats the same experiences as Figure 5 in the main text, but with the addition of LAMP instead of l1-based pruning. The conclusions about the usefulness of our added rescaling and soft thresholding remain the same: rescaling is only useful when combined with soft-thresholding. LAMP doesn't show any additional performance increase aver traditional l1-based pruning while being more expensive FLOPS-wise. Although not shown on this graph, LAMP, combined with STE alone, has the advantage of not suffering from layer collapse due to its formal definition.
\subsection{Training parameters}
All training configuration files and precise network structures are uploaded made open-source at \url{https://github.com/vanderschuea/stthree}. A quick rundown is provided in Table \ref{tab:hyper} for easy comparison.
\subsubsection{Hyper-parameters}
The hyper-parameters used for our training of the networks are presented in Table \ref{tab:hyper}. All results on Cifar-10(0) are reported from a train/val/test split of sizes 45k/5k/10k. The results on ImageNet are reported on the validation set.

\begin{table*}
\centering
\begin{tabular}{||r | c  |  c ||} 
 \hline
 Model & ResNet-20/VGG-11/WideResNet-34x2 & ResNet-50/MobileNetv1 \\ 
 \hline\hline
 epochs & 160 & 100\\\hline
 optimizer & \multicolumn{2}{c||}{SGD}\\\hline
 lr & 0.1 & 0.2\\\hline
 momentum & \multicolumn{2}{c||}{0.9}\\\hline
 batch-size & 128 & 256\\\hline
 weight-decay & 1e-4 & see below\\\hline
 lr-decay & Step-lr@[80,120] & cosine-lr (+ warmup)\\\hline
 grad. clip. & \multicolumn{2}{c||}{3.0}\\
 
 \hline

\end{tabular}

\caption{
Hyper-parameters used for training. (left: networks on Cifar-10, right: networks on ImageNet)}\label{tab:hyper}

\end{table*}

Due to the double regularization-effect of pruning and weight-decay, the weight-decay parameter is slowly decreased as follows at different sparsity ratios:
\begin{itemize}
\item ResNet50: 1e-4 (80-95), 1e-5 (96.5-98), 0.0 (99)
\item MobileNetv1: 1e-4 (80), 1e-5 (89-95)
\end{itemize}
We do not apply decaying on the weight-decay parameter for the cifar-10(0) models as they don't benefit much from such optimizations. 

\subsubsection{Augmentations}
\paragraph{Cifar-10(0)} The standard 4x4 padding is applied before a 32x32 crop is taken. The data is normalized and also randomly horizontally flipped.
\paragraph{ImageNet}  Images are randomly cropped and resized to 224x224. The data is normalized and also randomly horizontally flipped. (Test images are resized to 256x256 and a centercrop of size 224x224 is taken for evaluation)

\subsection{Pseudo-code}

\alglanguage{pseudocode}
\begin{algorithm}[h]
\small
\caption{PyTorch-like Pseudo-code for ST-3\colorbox{pink}{$^\sigma$} without and \colorbox{pink}{with Sparsity Distribution Suggestion}}
\label{alg:pseudo}
\begin{algorithmic}[1]   
\For {$i = 1 \to \text{n\_epochs}$}
	\For {$batch = 1 \to \text{len(train\_loader)}$}
		\State sp\_ratio = \text{get\_sp\_ratio}(i, batch)
    	\Comment update sparsity ratio
    	\State weights = torch.tensor([])
		\Comment aggregate all weights    	
    	\For {module in model.modules}
    		\State w = module.weight
    		\State weights.append(w.flatten() \colorbox{pink}{* $\sqrt{\text{w.shape.sum()}}$})
    	\EndFor
		\State th = weights.abs().quantile(sp\_ratio)
		\State W$_{\text{sparse}}$ = \{\}
		\Comment aggregate sparsified weights
    	\For {name, module in model.named\_modules}
    		\State mth = th \colorbox{pink}{/ $\sqrt{\text{w.shape.sum()}}$}
    		\State \textbf{with} STE():
			\Comment straight-through gradients
			\Indent
				\State w = module.weight
				\Comment dense weights
				\State w$_{\text{sparse}}$ = sign(w) * ($\|\text{w}\|$-mth).clip(0)
				\Comment soft-th
				\State w$_{\text{sparse}}$ *= get\_scale(w, mth)
				\Comment apply rescaling
    			\State W$_{\text{sparse}}[\text{name}]$ = w$_{\text{sparse}}$
    		\EndIndent
    	\EndFor    	
    	
    	\State inputs = trainloader[batch]
    	\State loss = model(W$_{\text{sparse}}$, inputs)
    	\State loss.backward()
    	\Comment + update all dense model weights
    \EndFor
\EndFor
\Statex
\end{algorithmic}
  \vspace{-0.4cm}%
\end{algorithm}
\newpage
\subsection{Table of results}
Table \ref{tab:mobile} and \ref{tab:resnet} contain the exact numerical values used for the graphs presenting the results on ImageNet. The values for the other methods, are taken from their respective papers which use the same training times and data-augmentations. No 'best' values are put in bold as it would depend on whether comparison is made on Accuracy or GFLOPS as the common denominator. For a better understanding, we refer to the figures presented in the paper's main body.

\begin{table}
  \begin{center}
    {\small{
\begin{tabular}{l|llr}
\toprule
Method & Accuracy $\left[\%\right]$ & Sparsity $\left[\%\right]$ & GFLOPS \\
\midrule
GMP & 67.7 & 74.11 & 163\\
STR & 68.35 & 75.28 & 101\\
STR & 66.52 & 79.07 & 81\\
RigL+ERK & 68.39 & 75 & 296 \\
ST-3$^\sigma$ (Ours) &  68.67 & 80 & 135\\
ST-3 (Ours) & 69.1 & 80 & 181\\
\midrule
STR & 64.83 & 85.8 & 55 \\
GMP & 61.8 & 89.03 & 82\\
STR & 62.1 & 89.01 & 42\\
STR & 61.51 & 89.62 & 40\\
RigL+ERK & 62.33 & 90 & 154\\
ProbMask & 65.19 & 89 & 63\\
ST-3$^\sigma$ (Ours) & 66.15 & 89 & 86 \\
ST-3 (Ours) & 66.67 & 89 & 119\\
\midrule
RigL+ERK & 56.27 & 95 & 103\\
ProbMask & 60.1 & 94.1 & 34\\
ST-3$^\sigma$ (Ours) & 60.94 & 94.1 & 52\\
ST-3 (Ours) & 61.19 & 94.1 & 74\\

\bottomrule
\end{tabular}
}}
\end{center}
\caption{Numerical Values for the figures with results on ImageNet w/ MobileNetv1}\label{tab:mobile}
\end{table}
\begin{table*}
  \begin{center}
    {\small{
\begin{tabular}{l|llr}
\toprule
Method & Accuracy $\left[\%\right]$ & Sparsity $\left[\%\right]$ & GFLOPS \\
\midrule
RigL & 74.6 & 80 & 940 \\
RigL+ERK & 75.1 & 80 & 1717 \\
DNW & 76 & 80 & 818 \\
GMP & 75.6 & 80 & 818 \\
STR & 76.19 & 79.55 & 766\\
STR & 76.12 & 81.27 & 705 \\
GraNet & 76 & 80 & 1431 \\
ST-3$^\sigma$ (Ours) & 76.44 & 80 & 739\\
ST-3 (Ours) & 76.95 & 80 & 1215\\
\midrule
RigL & 72.0 & 90 & 531 \\
RigL+ERK & 73.0 & 90 & 981\\
DNW & 74 & 90 & 409 \\
GMP & 73.91 & 90 & 409\\
STR & 74.73 & 87.7 & 402 \\
STR & 74.31 & 90.23 & 343 \\
STR & 74.01 & 90.55 & 341 \\
GraNet & 74.5 & 90 & 818 \\
ProbMask & 74.68 & 90 & 381\\
ST-3$^\sigma$ (Ours) & 75.28 & 90 & 397 \\
ST-3 (Ours) & 76.03 & 90 & 764 \\
\midrule
RigL & 67.5 & 95 & 327 \\
RigL+ERK & 70 & 95 & 491 \\
DNW & 68.3 & 95 & 204 \\
GMP & 70.59 & 95 & 204 \\
STR & 70.97 & 94.8 & 182 \\
STR & 70.4 & 95.03 & 159\\
STR & 70.21 & 95.15 & 162\\
ProbMask & 71.5 & 95 & 205 \\
ST-3$^\sigma$ (Ours) & 73.69 & 95 & 219\\
ST-3 (Ours) & 74.46 & 95 & 436\\
\midrule
RigL &  64.5 & 96.5 & 164\\
RigL+ERK & 67.2 & 96.5 & 286 \\
STR & 67.78 & 96.11 & 127 \\
STR & 67.22 & 96.53 & 117\\
ST-3$^\sigma$ (Ours) & 72.62 & 96.5 & 167 \\
ST-3 (Ours) & 73.31 & 96.5 & 351 \\
\midrule
DNW & 58.2 & 98 & 82\\
GMP & 57.9 & 98 & 82\\
STR & 62.84 & 97.78 & 80\\
STR & 61.46 & 98.05 & 73\\
STR & 59.76 & 98.22 & 64\\
ProbMask & 66.83 & 98 & 105\\
ST-3$^\sigma$ (Ours) & 69.75 & 98 & 116 \\
ST-3 (Ours) & 70.46 & 98 & 220 \\
\midrule
GMP & 44.78 & 99 & 41 \\
STR & 54.79 & 98.79 & 54 \\
STR & 51.82 & 98.98 & 47\\
STR & 50.35 & 99.1 & 44\\
ProbMask & 61.07 & 99 & 55\\
ST-3$^\sigma$ (Ours) & 63.25 & 99 & 69\\
ST-3 (Ours) & 63.88 & 99 & 120\\

\bottomrule
\end{tabular}
}}
\end{center}
\caption{Numerical Values for the figures with results on ImageNet w/ ResNet-50}\label{tab:resnet}
\end{table*}

\end{document}